Short Paper*
# Computer-Generated Sand Mixtures and Sand-based Images

Ryan A. Subong
Center of Research and Development, Western Institute of Technology, Philippines
crd@wit.edu.ph
(corresponding author)

Alma Jean D. Subong
School of ICT, West Visayas State University, Philippines
ajsubong@wvsu.edu.ph

## Abstract

*Purpose* – This paper aims to verify the effectiveness of the software implementation of the proposed algorithm in creating computer-generated images of sand mixtures using a photograph of sand as an input and its effectiveness in converting digital pictures into sand-based images out of the mixtures it generated.

*Method* – Visually compare the photographed image of the actual mixtures to its computer-generated counterpart to verify if the mixture generation produces results as expected and compare the computer-generated sand-based images with its source to verify image reproduction maintains same image content.

*Results* – The mixture comparison shows that the actual and the computer-generated ones have similar overall shade and color. Still, the generated one has a rougher texture and higher contrast due to the method of inheriting visual features by pixel, not by individual sand particles. The comparison of the sand-based image and its source has demonstrated the software's ability to maintain the essence of its contents during conversion while replacing its texture with the visual properties of the generated sand mixture.

*Conclusion* – The software implementation of the proposed algorithm can effectively use the images of sand to generate images of its mixtures and use those mixture images to convert a digital picture into a computer-generated sand-based image.

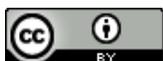

*Recommendations* – Explore the possibility of revising the method of inheriting visual features of sands by individual sand particles in generating images of sand mixtures instead of inheriting by pixel to improve the accuracy of the image generation.

*Research Implications* – Automating the sand ratio generation eliminates some of the manual processes involved in creating sand-based artworks, increasing their production efficiency.

*Keywords* – image processing, computer-generated images, disperse-dot, sand art

## INTRODUCTION

Since 2019, the corresponding author of this paper has been exploring an innovative method involving using natural, untreated sand of various colors to duplicate human portrait photographs. This process involves mapping the source image into different shaded areas, redrawing the shade map onto a target board, applying glue, and then sprinkling sand mixtures onto the glued areas to match the overall tone. Figure 1 showcases one of the author's early works: a 30x50 cm sand-based artwork depicting Miss Universe Philippines 2020, Ms. Rabiya Mateo, utilizing black, beige, and white coarse sand.

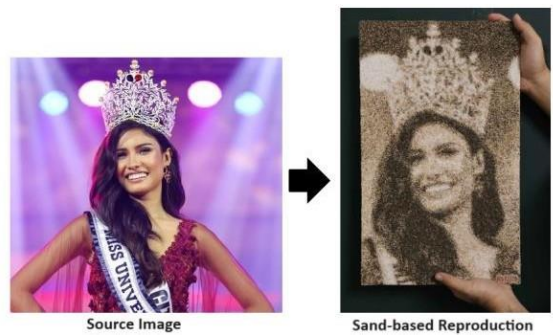

*Figure 1.* Conversion of image

In this imaging process, the sand as the medium of creating images is intended to be used as it is without the introduction of dyes, paints, heat, or any factor that might influence its original and natural color. To emulate the varying shades in an image, sands of lighter and darker shades are mixed in a ratio wherein its overall mean gray value will fall somewhere between the given lighter and darker sand. Since the individual particles of the sand are fixed in size, the shade control will be done similarly to how the halftoning of the disperse-dot method works. Shown in Figure 2 is how the varying ratio of the sands within the mixture visually affects the shade of the various areas of the image.

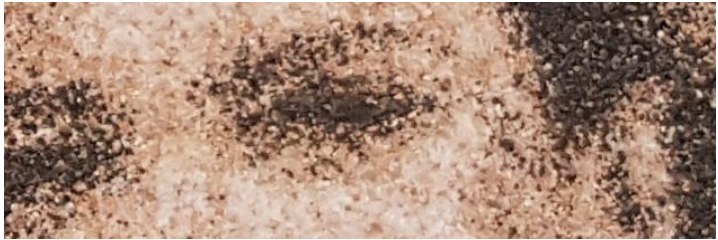
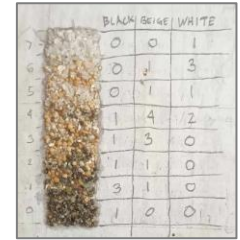

*Figure 2.* Close-up view of the sand-based image      *Figure 3.* Early mixture set

The said imaging process starts with the preparation of the sand mixture set using two or more uniformed colored sands of various colors, preferably one should be very light in color and the other is very dark to widen the range of colors available for the image reproduction. A sand mixture set is a set of sand mixtures wherein each mixture is composed of one or more sands in a defined ratio which when mixed creates an overall shade intensity unique to each as shown in Figure 3. In the early works of the author, the sand mixture sets are created manually through repetitive mixing experiments and visual inspection processes for each mixture until a satisfactory overall mixture tone or shade is achieved. Once all the mixtures of the set are complete, only then the author can proceed with the creation of the target image. This trial-and-error method of coming up with the mixture set is however a time-consuming process that must be repeated if a new image requires a new set of sand or a different number of colors. If only the generation of the sand ratio of the mixture set is automated with a visual simulation of its result, the whole sand images process could significantly be done in less time.

This paper aims to create an algorithm for a computer program that will evaluate the color of the various sands to be used for imaging and generate a set of mixing ratio formulas from the possible lightest color it could make up from the given sands, up to its possible darkest color. The program will also generate a visual simulation of each of the mixtures and a simulated preview of what will a digital image look like if it is redrawn using mixtures as its coloring medium.

## LITERATURE REVIEW

Sand is a product of thousands of years of rock erosion and has been used as a building and construction material since ancient times (Dan Gavriletea, 2017). Sand has been used in the creation of various flat or two-dimensional artworks such as sand drawings (Zlatev et al., 2023) and sand paintings (Wu & Conti, 2015) dating back centuries by different civilizations around the world for both cultural and creative purposes. These artworks are however only capable of generating images out of simple lines and patterns.

To give a better understanding of how the imaging process of this paper incorporates shading techniques to create more complex images using sand, this literature review aims to discuss the existing imaging and printing technologies that are involved in its process. The process in this paper evaluates images as digital grayscale images, wherein each pixel carries only its gray intensity value but none of the other colors (Biswas et al., 2011). The

3121

process chooses the grayscale version of the images instead of its originally colored format as it simplifies and reduces the computational process and requirements it must undergo (Kanan & Cottrell, 2012). To convert colored images into grayscale, the average method is used wherein the grayscale value of each pixel is directly from the calculated average of that pixel's red, green, and blue channels (Wan & Xie, 2016).

Grayscale images in this paper are quantified through the computation of its mean gray value, wherein the intensity of all the pixels comprising the image is measured and averaged to gauge the overall brightness or darkness of the entire image (Maurer & Stapleton, 2021). An 8-bit grayscale image will have a mean gray value ranging from 0 to 255, wherein 0 is black or the darkest and 255 is white or the lightest (Saravanan, 2010).

Continuous-tone color images can be generated using only a limited number of colors through digital color halftoning (Baqai et al., 2005). Halftoning achieves this by simulating a range of shades by adjusting the density or arrangement of dots or patterns that comprise the image (Ulichney, 1999). Since this method is used in conventional computer printers, wherein the inks are limited only to the colors cyan, yellow, magenta, and black (Choi et al., 2020), the same can be used when using sands of limited colors.

In halftoning monochromatic pictures, images can be produced either by cluster-dot or disperse-dot method (Huang et al., 2018). The cluster-dot method arranges dark dots in a grid-like pattern against a light background and produces various gray levels by modulating the size of those dots (Goyal et al., 2013). The bigger the dots, the darker the area it occupies. Conversely, disperse-dot produces gray levels by varying the number of similarly sized dark dots arranged in a light-colored background. The more crowded the dots, the darker the area they occupy (Gooran & Kruse, 2015). Since the sizes of sand particles are already fixed, the disperse-dot method will be used in image generation like how impact printers such as dot-matrix printers can print images by imprinting an extensive array of fixed-size black dots on paper (Harris, 1984).

## METHODOLOGY

This paper proposes an algorithm that will evaluate the visual properties of a set of sand using its photographed image and then indicate recommending sand mixing ratios in creating its mixture set. The algorithm is then implemented into a computer software using VB.net programming language that reads the digital images of the sands and the image for reproduction to generate a picture of what the individual mixture looks like after it has been mixed and what the image subject for reproduction will look like it if it is turned into a sand-based artwork. Figure 4 shows the sequence of verification done for this study.

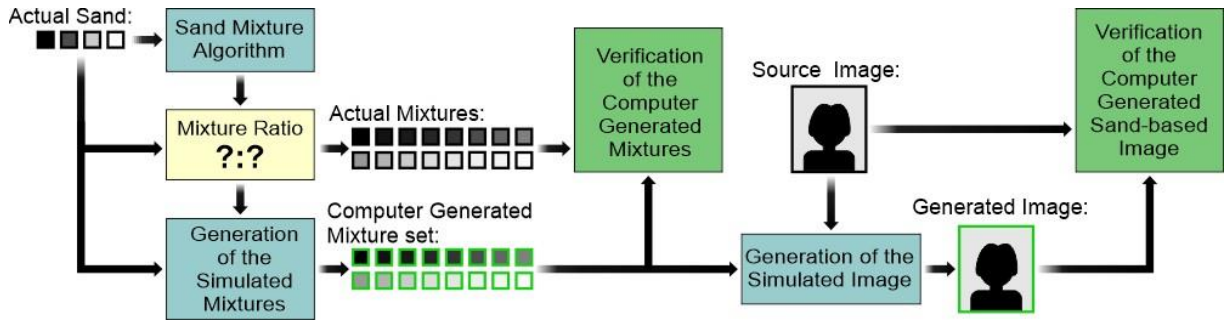

*Figure 4.* Verification Sequence

All photographed images of sand and its mixture sets in this paper were taken using an OPPO Reno8 T camera phone with a manual pro setting of ISO (light sensitivity) = 125, shutter = 1/25, EV (exposure value) = 0.0, and WB (white balance) = 4800. They were taken inside a lightbox with a camera-to-sample distance of 180 mm and a maintained white light illuminance of 160 lx.

## *Nature of the Algorithm*

The core process of the algorithm is how to create mixtures of sand that emulate how the disperse-dot halftoning method creates a gradual shade transition from darkest to lightest, but instead of controlling the number of imprinted dark dots over a light background, the algorithm controls the quantity of darker of sands over the lighter ones within the mixture. To bridge the shade transition between two areas of sand, the algorithm has the first sand started in its pure form and then increasingly contaminates it with the second sand for each succeeding mixture while decreasing the amount of the first sand. This will be continued until the last mixture no longer contains the first sand, leaving only the second sand in its pure form. Shown in Table 1 is the ratio mixture set of Sand A and B if the two sands are to be bridged by three mixtures in between.

Table 1. Sand A to B Bridging Mixture Set

| Sand | Ratio 1 | Ratio 2 | Ratio 3 | Ratio 4 | Ratio 5 |
|---|---|---|---|---|---|
| A | 1 (100%) | 3 (75%) | 2 (50%) | 1 (25%) | 0 (0%) |
| B | 0 (0%) | 1 (25%) | 2 (50%) | 3 (75%) | 1 (100%) |

For example, the mixture set involves the computation of four different kinds of sand; the actual mean gray value of each given sand is first computed from their corresponding photographed images to use its values to arrange the sands from darkest to lightest shade tagged as Sand A, B, C and D respectively. Aiming for the fixed 16-mixture set, the darkest Sand A in its pure form will be assigned as the 1st mixture, and the lightest Sand D will be the 16th. The mean gray value difference between Sand A and D will be then divided into 14 other expected mean gray values in equal intervals wherein the mixture number of the remaining Sand B and C are assigned wherever their mean gray values are nearest to. Since all Sand A, B, C, and D are assigned with their mixture assignment at this point, the proposed algorithm can be applied to bridge the mixture shade between Sand A and B, B, and C, and then C and D. The entire steps of this process are shown at Figure 5.

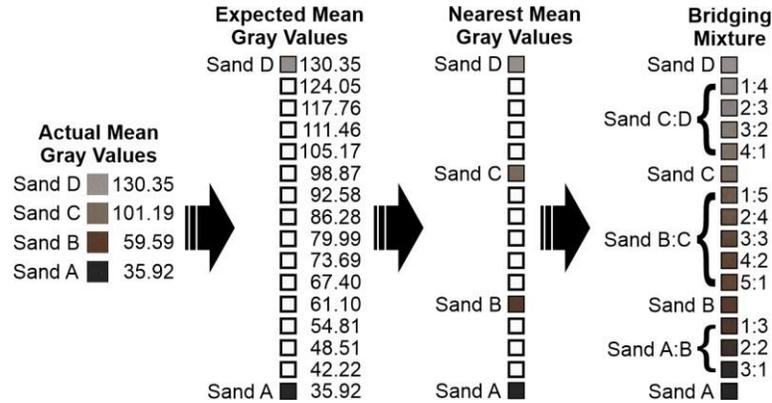

*Figure 5.* Algorithm implementation on 4-sand mixture set

The software implementation of the algorithm uses only the digital images of the involved sands as its inputs to automatically compute the sand ratio of the mixtures. Shown in Figure 6 is the input interface of the software where it uses the actual photographed images of the involved sands as its input, and its resulting output interface where the ratio parts and percentage of the sands per mixture are indicated after its computation. Also, in the same output interface, the image simulation of the computer-generated mixture is shown, which will be discussed next in this paper.

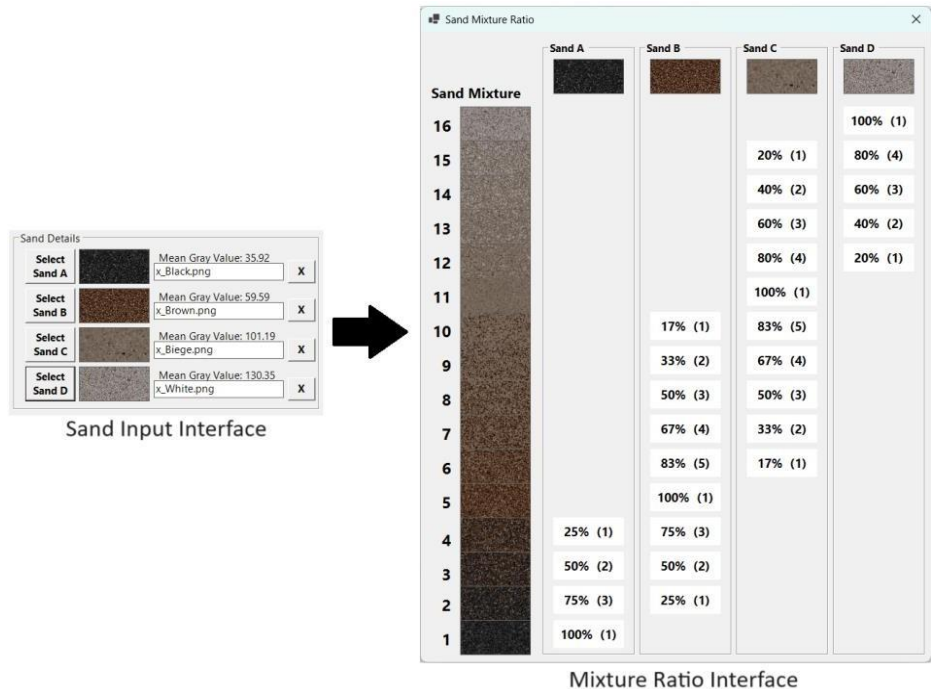

*Figure 6.* Sand Mixture Ratio Software Interface

3124

### Verification of the Computer-Generated Mixture

To generate a preview of a mixture, the software first obtains the photographed image of the two involved sands and allocates a blank pixel image where the simulated mixture will be drawn. All pixels of the blank image are then colored depending on their position on the blank image, like in the alternating pattern of a chessboard. First, all pixels that fall within the positions of the black squares in the pattern will randomly pick their sand with the possibility of being picked influenced by their parts ratio. For example, if the mixture consists of Sand A and B with a ratio of 3:1, there will be a 75% chance that Sand A will be picked and a 25% chance that Sand B will be picked. The software will then pick a random pixel out of the chosen sand and finally inherit its pixel color to the corresponding pixel of the blank image. This random pixel-picking process will simulate the result of the mechanical mixing of two different areas of sand in defined quantities.

After that, all remaining pixels that fall within the positions of the white squares in the pattern will compute the average color of all adjacent pixels belonging to the black square positions and then use that average to color themselves. This will soften the color transition between pixels taken from the sand and complete the whole preview image. Figure 7 shows the pattern of pixel assignments used to generate the preview image.

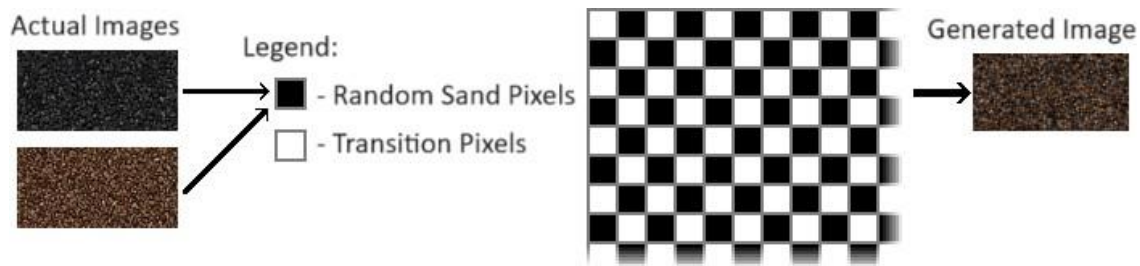

*Figure 7.* Mixture simulation pixel assignments

To verify the software's effectiveness in simulating the mixture's result, the preview image generated by the 4-sand mixture set will be visually compared to the photographed image of its actual counterparts. Suppose the entire set of the generated mixtures continuously emulates the general color transition of the set of its counterpart from darkest to lightest. In that case, the software is deemed acceptable in generating images of simulated sand mixtures for preview purposes.

### Verification of the Computer-Generated Sand-based Image

The other intended feature of the software is the generation of how a digital image will look if its colors are replaced with sand mixtures. This will give users a preview of the end-product image before any mixing activity begins.

First, the software is provided with the photographed images of the sands to be used. It will then use those images to calculate the mixture ratio of its 16-mixture set and generate the simulated image of each mixture. The software then evenly assigns each mixture of the set to a corresponding range of the 256-color grayscale spectrum, as shown in Table 2, so it would know what mixture set will be used for each pixel grayscale value. However, the range of assignments can be manually adjusted to enhance the final image.

Table 2. Mixture Set Assignment

| Assigned Mixture Number | Grayscale Range | Assigned Mixture Number | Grayscale Range |
|---|---|---|---|
| 1 | 0 – 15 | 9 | 128 – 143 |
| 2 | 16 – 31 | 10 | 144 – 159 |
| 3 | 32 – 47 | 11 | 160 – 175 |
| 4 | 48 – 63 | 12 | 176 – 191 |
| 5 | 64 – 79 | 13 | 192 – 207 |
| 6 | 80 – 95 | 14 | 208 – 223 |
| 7 | 96 – 111 | 15 | 224 – 239 |
| 8 | 112 – 127 | 16 | 240 – 255 |

The software then provides the source image to be reproduced. It will internally convert it into its 8-bit or 256-color grayscale version so it can use each pixel's grayscale value for the mixture assignment. The software will also enlarge the image so each of its pixels is now at least four pixels in size. This enlargement of the pixels will give the mixture image enough area to redraw the texture of the mixture and retain its visual properties if it is going to be drawn over that area.

Finally, the software will generate the preview of the sand-based image for reproduction by redrawing it using its pixel's grayscale values to determine what sand mixture image will be drawn over the enlarged pixels instead of its original grayscale color. The output is expected to exhibit the exact content of the source picture while having colors and textures inherited from the sand mixtures.

To verify the effectiveness of the software in its conversion of images into its sand-based previews, an 1890 head photo of Dr. José Rizal, a national hero of the Philippines, and the painting of Warner Sallman titled Head of Christ will be converted using a 4-sand mixture set. The generated previews will be compared visually against their source images. If the content of the source image can still be recognizable from its sand-based counterpart despite having a rougher texture and less color profile, the conversion process of the software is then deemed acceptable.

**RESULTS**

For the verification of the computer-generated mixture set, Table 3 shows a direct comparison of its simulated image of the 4-sand mixture set generated by the software of this study with the photographs of their corresponding actual mixtures that are manually mixed based on the recommendations of the generated mixture ratio.

Table 3. Comparison of mixture images

| Mixture # | 1 | 2 | 3 | 4 | 5 | 6 | 7 | 8 |
|---|---|---|---|---|---|---|---|---|
| Actual Photo | 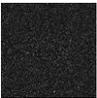 | 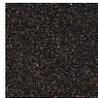 | 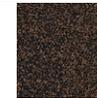 | 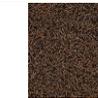 | 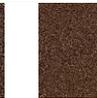 | 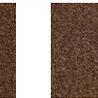 | 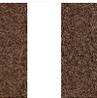 | 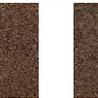 |
| Computer Generated | 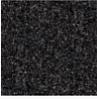 | 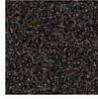 | 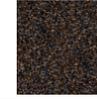 | 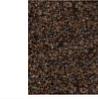 | 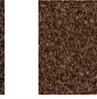 | 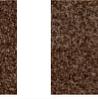 | 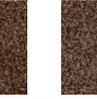 | 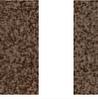 |
| Mixture # | 9 | 10 | 11 | 12 | 13 | 14 | 15 | 16 |
| Actual Photo | 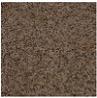 | 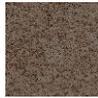 | 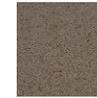 | 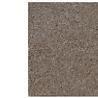 | 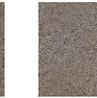 | 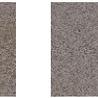 | 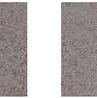 | 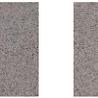 |
| Computer Generated | 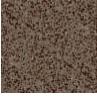 | 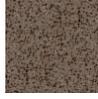 | 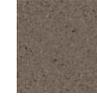 | 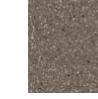 | 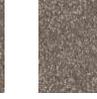 | 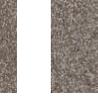 | 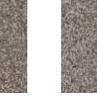 | 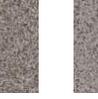 |

In verifying the computer-generated preview of the sand-based images, Figure 8 shows the two source images that are the subject of conversion and their sand-based version using the 4-sand mixture set.

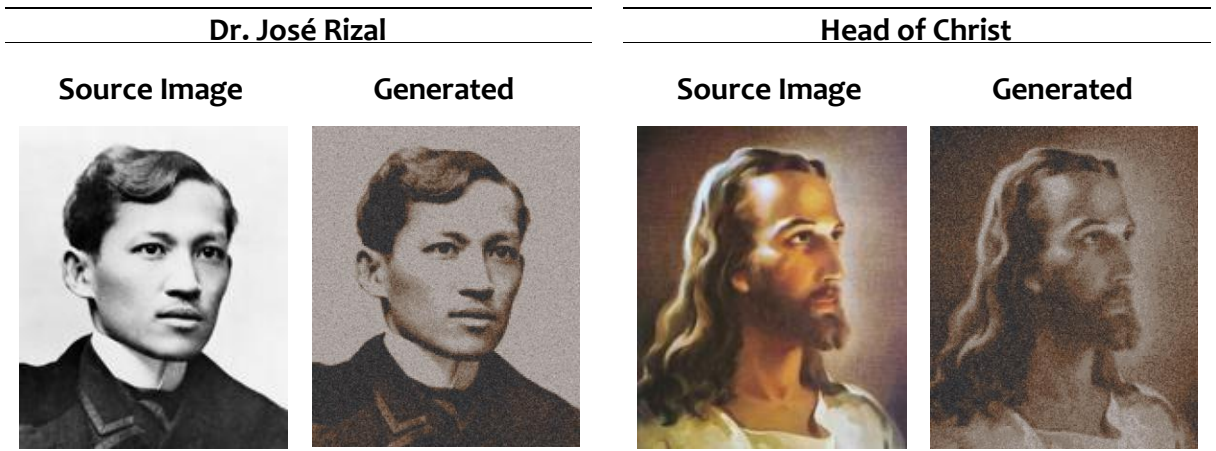

Figure 8. Comparison of Images

## DISCUSSION

When the actual photos of the mixtures are compared with their computer-generated counterpart, both visually show similar overall shade and color and transition from darkest to lightest. The generated image, however, is rougher in texture or higher than its corresponding actual photo as the extraction of features from the actual photo is picked by individual random pixels, not the entire single particle of the sand. This lack of particle features transferred from actual to generated results in a sudden switch of shades and colors between neighboring pixels, contributing to the sharp image contrast. Though this method does not accurately simulate what neighboring sand particles look like in an actual photograph, its generation of images has managed to have its overall shades and color identical to its actual counterpart. In verifying the computer-generated preview of the sand-based pictures, Figure 8 shows the two source images that are the subject of conversion and their sand-based version using the 4-sand mixture set.

The conversion of both the photo of Dr. José Rizal and the painting Head of Christ into a sand-based preview successfully redrawn the source with a sand-like texture and color while preserving the essence of its content. Even with the limited shade of 16 colors only, most details of the source image are still retained and recognizable.

## CONCLUSIONS AND RECOMMENDATIONS

The software implementation of the algorithm has been verified to effectively generate simulated previews of sand mixtures and sand-based images using only the images of the sands involved and the image that is subject to conversion. This successful verification justifies using the software as a mixing tool in creating sand-based image reproduction artworks.

Since the image of the computer-generated mixtures that are used to draw the sand-based image does not accurately duplicate what the actual sand mixture looks like at a very close view, it is recommended that another method should be explored where instead of inheriting the colors of the sand image per pixel for the mixture generation, the new process should inherit the entire visual features of the individual sand particle instead. Theoretically, this should mitigate the increasing image contrast of the resulting generated mixture and create more realistic previews of the sand-based images.

## IMPLICATIONS

This paper will increase the efficiency of the production of sand-based artwork by automating the creation of the sand mixture set for the craft and allowing users to use the computer-generated previews to make advanced adjustments with the sands even before the actual mixing of the sands begins.

## ACKNOWLEDGEMENT

The authors of this paper would like to thank the administrations of the Western Institute of Technology and West Visayas State University for supporting this project right from its proposal to up to its approval and execution.

## FUNDING

The study is funded by the Department of Science and Technology—Philippine

Council for Industry, Energy and Emerging Technology Research and Development (DOST-PCIEERD) under their Grants-in-Aid Program Call for Proposals for CY2023.

## DECLARATIONS

### *Conflict of Interest*

All authors declare that they have no conflicts of interest.

### *Informed Consent*

The image of the Head of Christ, painted by Warner Sallman, is a copyrighted property of Warner Press, Inc., Anderson, Indiana, USA. The author of this paper has been given emailed consent to use it by the Vice President of Production Management of Warner Press, Inc., with the condition that it be used for research purposes only, not for resell or profit.

### *Ethics Approval*

Ethics approval does not apply to this paper as the research did not gather data from human participants, animals, or vulnerable sources.

## Author's Biography

Dr. Ryan A. Subong is currently the Director of the Center of Research and Development of Western Institute of Technology, Iloilo City. He is a proud graduate of the Doctor of Engineering with specialization in Computer Engineering program of Technological Institute of the Philippines, Quezon City. He uses his time to turn his ideas, hobbies, and innovations into implemented projects with real-world applications and impact.

Dr. Alma Jean D. Subong is a passionate leader in educational technology with over 20 years of teaching experience. Leveraging her master's degree in IT and PhD in Education, Dr. Subong has spearheaded advancements in this field at West Visayas State University Janiuay Campus. Her dedication extends beyond the classroom, driving impactful contributions to the field of educational technology.